\documentclass[conference]{IEEEtran}
\IEEEoverridecommandlockouts
\usepackage{cite}
\usepackage{amsmath,amssymb,amsfonts}
\usepackage{algorithmic}
\usepackage{graphicx}
\usepackage{textcomp}
\usepackage{xcolor}

\usepackage{mathabx}

\def\BibTeX{{\rm B\kern-.05em{\sc i\kern-.025em b}\kern-.08em
    T\kern-.1667em\lower.7ex\hbox{E}\kern-.125emX}}
\begin{document}

\title{Planning with Learned Binarized Neural Networks Benchmarks for MaxSAT Evaluation 2021
}

\makeatletter
\newcommand{\linebreakand}{%
  \end{@IEEEauthorhalign}
  \hfill\mbox{}\par
  \mbox{}\hfill\begin{@IEEEauthorhalign}
}
\makeatother

\author{
\IEEEauthorblockN{Buser Say}
\IEEEauthorblockA{\textit{Monash University}\\
Melbourne, Australia \\
buser.say@monash.edu}

\and

\IEEEauthorblockN{Scott Sanner}
\IEEEauthorblockA{\textit{University of Toronto}\\
Toronto, Canada \\
ssanner@mie.utoronto.ca}

\and

\IEEEauthorblockN{Jo Devriendt}
\IEEEauthorblockA{\textit{KU Leuven}\\
Leuven, Belgium \\
jo.devriendt@kuleuven.be}

\linebreakand

\IEEEauthorblockN{Jakob Nordström}
\IEEEauthorblockA{\textit{University of Copenhagen}\\
Copenhagen, Denmark \\
jn@di.ku.dk}

\and

\IEEEauthorblockN{Peter J. Stuckey}
\IEEEauthorblockA{\textit{Monash University}\\
Melbourne, Australia \\
peter.stuckey@monash.edu}

}

\maketitle

\begin{abstract}
This document provides a brief introduction to learned automated 
planning problem where the state transition function is in the 
form of a binarized neural network (BNN), presents a general MaxSAT 
encoding for this problem, and describes the four domains, namely: 
Navigation, Inventory Control, System Administrator and Cellda, that 
are submitted as benchmarks for MaxSAT Evaluation 2021.
\end{abstract}


\section{Introduction}

Automated planning studies the reasoning side of acting in Artificial 
Intelligence, and automates the selection and ordering of actions to 
reach desired states of the world as best as possible~\cite{Nau2004}. 
An automated planning problem represents dynamics of the real-world 
using a model, which can either be manually 
encoded~\cite{Kautz1992,Hoffmann2001,Helmert2006,Pommerening2014,Davies2015}, 
or learned from data~\cite{Shen1989,Gil1992,Bennett1996,Benson1997}. 
In this document, we focus on the latter.

Automated planning with deep neural network (DNN) learned state transition 
functions is a two stage data-driven framework for learning and solving 
automated planning problems with unknown state transition 
functions~\cite{Say2017,Say2018,Say2020c}. The first stage of the framework 
learns the unknown state transition function from data as a DNN. The second 
stage of the framework plans optimally with respect to the learned DNN by 
solving an equivalent optimization problem (e.g., a mixed-integer programming 
model~\cite{Say2017,Say2019,Wu2020,Say2021}, a 0--1 integer programming 
model~\cite{Say2018,Say2020a}, a weighted partial MaxSAT 
model~\cite{Say2018,Say2020a}, a constraint programming 
model~\cite{Say2020b}, or a pseudo-Boolean optimization model~\cite{Say2020b}). 
In this document, we focus on the second stage of the data-driven framework 
where the learned DNN is a binarized neural network (BNN)~\cite{Hubara2016}.

The remaining of the document is organized as follows. We begin with the 
description of the learned automated planning problem and the binarized 
neural network (BNN). Then we present the weighted partial MaxSAT model 
of the general learned automated planning problem, and conclude with the 
description of four learned automated planning domains, namely: Navigation, 
Inventory Control, System Administrator and Cellda, that are submitted as 
benchmarks for MaxSAT Evaluation 2021.

\section{Automated Planning with Learned Binarized Neural Network 
State Transitions}

\subsection{Problem Definition}

\newcommand{\nn}{n} 
\newcommand{\mm}{m} 

A fixed-horizon learned deterministic automated planning 
problem~\cite{Say2017,Say2018,Say2020b} is a tuple 
$\tilde{\Pi} = \langle S,A,C,\tilde{T},V,G,R,H \rangle$, where $S=\{s_1, 
\dots ,s_{\nn}\}$ and $A=\{a_1, \dots ,a_{\mm}\}$ are sets of state and 
action variables for positive integers $\nn, \mm \in \mathbb{Z^{+}}$ with 
domains $D_{s_1}, \dots, D_{s_{\nn}}$ and $D_{a_1}, \dots, D_{a_{\mm}}$ 
respectively. Moreover, 
$C: 
D_{s_1} \times \dots \times D_{s_{\nn}} \times D_{a_1} \times \dots \times D_{a_{\mm}} 
\rightarrow \{\mathit{true},\mathit{false}\}$ is the global function, 
$\tilde{T}: 
D_{s_1} \times \dots \times D_{s_{\nn }} \times D_{a_1} \times \dots \times D_{a_{\mm}} 
\rightarrow 
D_{s_1} \times \dots \times D_{s_{\nn }}
$ is the learned state 
transition function, and $R: 
D_{s_1} \times \dots \times D_{s_{\nn }} \times D_{a_1} \times \dots \times D_{a_{\mm}} 
\rightarrow \mathbb{R}$ is the reward 
function. Finally, $V$ is a tuple of constants $\langle V_1,\dots, V_{\nn } 
\rangle \in D_{s_1} \times \dots \times D_{s_{\nn }}$ 
denoting the initial values of all state variables, 
$G: D_{s_1} \times \dots \times D_{s_{\nn }}  \rightarrow \{\mathit{true},\mathit{false}\}$ 
is the goal state function, and $H\in \mathbb{Z}^{+}$ is the planning horizon.

A \emph{solution} to (i.e., a \emph{plan} for) 
$\tilde{\Pi}$ is a tuple of values $\bar{A}^{t} = \langle 
\bar{a}^{t}_{1}, \dots, \bar{a}^{t}_{\mm} \rangle \in D_{a_1} \times \dots 
\times D_{a_{\mm}}$ for all action variables $A$ over time steps 
$t\in \{1,\dots, H\}$ such that $\tilde{T}(\langle \bar{s}^{t}_{1}, \dots, 
\bar{s}^{t}_{\nn }, \bar{a}^{t}_{1}, \dots, \bar{a}^{t}_{\mm} \rangle) = 
\langle \bar{s}^{t+1}_{1}, \dots, \bar{s}^{t+1}_{\nn } \rangle$ and $C(\langle 
\bar{s}^{t}_{1}, \dots, \bar{s}^{t}_{\nn }, \bar{a}^{t}_{1}, \dots, 
\bar{a}^{t}_{\mm} \rangle) = \mathit{true}$ holds for time steps 
$t\in \{1,\dots, H\}$, $V_i = \bar{s}^{1}_{i}$ for all $s_i \in S$ and 
$G(\langle \bar{s}^{H+1}_{1}, \dots, \bar{s}^{H+1}_{\nn } \rangle) = 
\mathit{true}$. It has been shown that finding a feasible solution to 
$\tilde{\Pi}$ is \textit{NP}-complete~\cite{Say2020b}.
An \emph{optimal solution} to (i.e., an \emph{optimal plan} for) $\tilde{\Pi}$ 
is a solution such that the total reward 
$\sum_{t = 1}^{H} R(\langle \bar{s}^{t+1}_{1}, \dots, \bar{s}^{t+1}_{\nn }, 
\bar{a}^{t}_{1}, \dots, \bar{a}^{t}_{\mm} \rangle)$ is maximized.

We assume that the domains of action and state variables are binary unless 
otherwise stated\footnote{When the domain of a variable is not binary (e.g., 
see Inventory Control), we can use the following approximation 
$x \approx (-2^{m_1 -1}x_{m_1} + \sum_{i=1}^{m_1 -1} 2^{i-1}x_{i}) 10^{m_2}$ 
for integers $m_1\in \mathbb{Z}^{+}$ and $m_2\in \mathbb{Z}$.}, 
the functions $C, G, R$ and function $\tilde{T}$ are known, 
functions $C, G$ can be equivalently represented by $J_{C}\in 
\mathbb{Z^{+}}$ and $J_{G}\in \mathbb{Z^{+}}$ linear 
constraints, function $R$ is a linear expression and function 
$\tilde{T}$ is a learned BNN~\cite{Hubara2016}.

\subsection{Binarized Neural Networks}

Binarized neural networks (BNNs) are DNNs with binarized weights and activation 
functions~\cite{Hubara2016}. Given $L$~layers with layer width~${W_l}$ of 
layer $l \in \{1,\dots,L\}$, and a set of neurons 
$J(l)=\{u_{1,l}, \dots ,u_{W_l,l}\}$, is stacked in the following order.

\paragraph{Input Layer} The first layer consists of neurons $u_{i,1}\in J(1)$ 
that represent the domain of the learned state transition function $\tilde{T}$ 
where neurons $u_{1,1}, \dots , u_{\nn ,1} \in J(1)$ represent the state variables 
$S$ and neurons $u_{\nn +1,1}, \dots ,u_{\nn +\mm,1} \in J(1)$ represent the 
action variables $A$. During the training of the BNN, values $0$ and $1$ of action 
and state variables are represented by $-1$ and $1$, respectively.

\paragraph{Batch Normalization Layers} For layers $l \in \{2,\dots,L\}$, 
\sloppy Batch Normalization~\cite{Ioffe2015} sets the weighted sum 
of outputs at layer $l-1$ in $\triangle_{j,l} = \sum_{i\in J(l-1)} w_{i,j,l} 
y_{i,l-1}$ to inputs $x_{j,l}$ of neurons $u_{j,l} \in J(l)$ using the formula  
$x_{j,l} = \frac{\triangle_{j,l} - \mu_{j,l}} {\sqrt[]{\sigma^{2}_{j,l} 
+ \epsilon_{j,l}}}\gamma_{j,l} + \beta_{j,l}$, 
where $y_{i,l-1}$ is the output of neuron $u_{i,l-1}\in J(l-1)$, 
and the parameters are the weight~$w_{i,j,l}$, input mean~$\mu_{j,l}$, input 
variance~$\sigma^{2}_{j,l}$, numerical stability constant~$\epsilon_{j,l}$, 
input scaling~$\gamma_{j,l}$, and input bias~$\beta_{j,l}$, all computed at 
training time.

\paragraph{Activation Layers} 
Given input $x_{j,l}$, the activation function $y_{j,l}$ computes the 
output of neuron $u_{j,l}\in J(l)$ at layer $l\in \{2,\dots,L\}$,
which is~$1$ if $x_{j,l}\geq 0$ and $-1$ otherwise. The last activation 
layer consists of neurons $u_{i,L}\in J(L)$ that represent the 
codomain of the learned state transition function~$\tilde{T}$ such that 
$u_{1,L}, \dots ,u_{\nn ,L} \in J(L)$ represent the state variables~$S$.

The BNN is trained to learn the function $\tilde{T}$ 
from data that consists of measurements on the domain and codomain of 
the \textit{unknown} state transition function $T: D_{s_1} \times \dots 
\times D_{s_{\nn }} \times D_{a_1} \times \dots \times D_{a_{\mm}} 
\rightarrow D_{s_1} \times \dots \times D_{s_{\nn }}$.

\section{The Weighted Partial MaxSAT Model}

In this section, we present the weighted partial MaxSAT 
model~\cite{Say2018,Say2020a} of the learned automated 
planning problem.

\subsection{Decision Variables} 

The weighted partial MaxSAT 
model uses the following decision variables:
\begin{itemize}
\item ${X}_{i,t}$ encodes whether action variable $a_i\in A$ 
is executed at time step $t\in \{1,\dots,H\}$ or not.
\item ${Y}_{i,t}$ encodes whether state variable $s_i\in S$ 
is true at time step $t\in \{1,\dots,H+1\}$ or not.
\item ${Z}_{i,l,t}$ encodes whether neuron $u_{i,l}\in J(l)$ 
in layer $l\in \{1,\dots,L\}$ is active at time step 
$t\in \{1,\dots,H\}$ or not.
\end{itemize}

\subsection{Parameters} 

The weighted partial MaxSAT model uses the following parameters:
\begin{itemize}
\item $\bar{w}_{i,j,l}$ is the value of the learned BNN weight 
between neuron $u_{i,l-1}\in J(l-1)$ and neuron $u_{j,l}\in J(l)$ 
in layer $l\in \{2,\dots,L\}$.
\item $B(j,l)$ is the value of the bias for neuron $u_{j,l}\in J(l)$ 
in layer $l\in \{2,\dots,L\}$. Given the values of learned parameters 
$\bar{\mu}_{j,l}$, $\bar{\sigma}^{2}_{j,l}$, $\bar{\epsilon}_{j,l}$, 
$\bar{\gamma}_{j,l}$ and $\bar{\beta}_{j,l}$, the bias is computed as 
$B(j,l) = \biggl \lceil \frac{\bar{\beta}_{j,l} 
\sqrt[]{\bar{\sigma}^{2}_{j,l} + \bar{\epsilon}_{j,l}}} 
{\bar{\gamma}_{j,l}} - \bar{\mu}_{j,l} \biggr \rceil$.
\item $r^{s}_{i}\in \mathbb{R}$ and $r^{a}_{i}\in \mathbb{R}$ are 
constants of the reward function $R$ that is in the form of 
$\sum_{i=1}^{\nn} r^{s}_{i} s_{i} + \sum_{i=1}^{\mm} r^{a}_{i} a_{i}$.
\item $c^{s}_{i,j}\in \mathbb{Z}$, $c^{a}_{i,j}\in \mathbb{Z}$ and 
$c^{k}_{j}\in \mathbb{Z}$ are constants 
of the set of linear constraints that represent the global function $C$ 
where each linear constraint $j\in \{1,\dots, J_{C}\}$ is in the form of 
$\sum_{i=1}^{\nn} c^{s}_{i} s_{i} + \sum_{i=1}^{\mm} c^{a}_{i} a_{i} 
\leq c^{k}_{j}$.
\item $g^{s}_{i,j}\in \mathbb{Z}$ and $g^{k}_{j}\in \mathbb{Z}$ are constants 
of the set of linear constraints that represent the goal state function 
$G$ where each linear constraint $j\in \{1,\dots, J_{G}\}$ is in the form 
of $\sum_{i=1}^{\nn} g^{s}_{i} s_{i} \leq g^{k}_{j}$.
\end{itemize}

\subsection{Hard Clauses}

The weighted partial MaxSAT model uses the following hard clauses.

\paragraph{Initial State Clauses}
The following conjunction of hard clauses sets the initial value $V_i$ of 
each state variable $s_i \in S$.
\begin{align}
&\bigwedge_{i=1}^{\nn}(\neg {Y}_{i,1} \vee V_{i}) \wedge 
({Y}_{i,1} \vee \neg V_{i}) \label{sat1}
\end{align}

\paragraph{Goal State Clauses}
The following conjunction of hard clauses encodes the set of linear constraints 
that represent the goal state function $G$.
\begin{align}
&\bigwedge_{j=1}^{J_{G}} 
Card(\sum_{i=1}^{\nn} g^{s}_{i} {Y}_{i,H+1} \leq g^{k}_{j})\label{sat2}
\end{align}
In the above notation, $Card$ produces the CNF encoding of a given linear constraint~\cite{Abio2014}.

\paragraph{Global Clauses}
The following conjunction of hard clauses encodes the set of linear constraints 
that represent the global function $C$.
\begin{align}
&\bigwedge_{j=1}^{J_{C}}\bigwedge_{t=1}^{H} 
Card(\sum_{i=1}^{\nn} c^{s}_{i} {Y}_{i,t} + \sum_{i=1}^{\mm} c^{a}_{i} {X}_{i,t} 
\leq c^{k}_{j})\label{sat3}
\end{align}

\paragraph{BNN Clauses}
The following conjunction of hard clauses maps the input and the output 
of the BNN onto the state and action variables.
\begin{align}
&\bigwedge_{i=1}^{\nn} \bigwedge_{t=1}^{H} (\neg{Y}_{i,t} \vee {Z}_{i,1,t}) \wedge ({Y}_{i,t} \vee \neg{Z}_{i,1,t})\label{sat4}\\
&\bigwedge_{i=1}^{\mm} \bigwedge_{t=1}^{H} (\neg{X}_{i,t} \vee {Z}_{i+\nn ,1,t}) \wedge ({X}_{i,t} \vee \neg{Z}_{i+\nn ,1,t})\label{sat5}\\
&\bigwedge_{i=1}^{\nn} \bigwedge_{t=1}^{H} (\neg{Y}_{i,t+1} \vee {Z}_{i,L,t}) \wedge ({Y}_{i,t+1} \vee \neg{Z}_{i,L,t})\label{sat6}
\end{align}

Finally, the following conjunction of hard clauses encodes the activation 
function of each neuron in the learned BNN.

\begin{align}
&\bigwedge_{l=2}^{L}\bigwedge_{u_{j,l} \in J(l)}\bigwedge_{t=1}^{H} Act\bigl(\nonumber\\
&(\sum_{u_{i,l-1}\in J(l-1)}{\bar{w}_{i,j,l}} (2
{Z}_{i,l-1,t} - 1) + B(j,l) \geq 0) = {Z}_{j,l,t}\bigr)\label{sat7}
\end{align}
In the above notation, $Act$ produces the CNF encoding of a given 
biconditional constraint~\cite{Say2020a} by extending the CNF 
encoding of Cardinality Networks~\cite{Asin2009}.

\subsection{Soft Clauses}

The weighted partial MaxSAT model uses the following soft clauses.

\paragraph{Reward Clauses}
The following conjunction of soft clauses encodes the reward function $R$.
\begin{align}
&\bigwedge_{t=1}^{H}\bigl(\bigwedge_{i=1}^{\nn} (r^{s}_{i}, {Y}_{i,t+1}) 
\wedge \bigwedge_{i=1}^{\mm} (r^{a}_{i}, {X}_{i,t})\bigr)\label{sat0}
\end{align}

\section{Benchmark Domain Descriptions}

In this section, we provide detailed description of four learned 
automated planning problems, namely: Navigation~\cite{Sanner2011}, 
Inventory Control~\cite{Mann2014}, System Administrator~\cite{Guestrin2001} 
and Cellda~\cite{Say2020a}.\footnote{The repository: https://github.com/saybuser/FD-SAT-Plan}

\begin{table}
  \centering
  \caption{The BNN architectures of all four learned 
automated planning problems.}
  \begin{tabular}{| l | c |}
    \hline
    Problem & BNN Structure\\
    \hline
    Discrete Navigation ($N=3$)& 13:36:36:9 \\ \hline
    Discrete Navigation ($N=4$)& 20:96:96:16 \\ \hline
    Discrete Navigation ($N=5$)& 29:128:128:25 \\ \hline
    Inventory Control ($N=2$)& 7:96:96:5 \\ \hline
    Inventory Control ($N=4$)& 8:128:128:5 \\ \hline
    System Administrator ($N=4$)& 16:128:128:12 \\ \hline
    System Administrator ($N=5$)& 20:128:128:128:15 \\ \hline
    Cellda (policy=x-axis)& 12:256:256:4 \\ \hline
    Cellda (policy=y-axis)& 12:256:256:4 \\ \hline
  \end{tabular}
  \label{tab:bnn}
\end{table}

\paragraph{Navigation} Navigation~\cite{Sanner2011} task for an 
agent in a two-dimensional ($N$-by-$N$ where $N\in \mathbb{Z}^{+}$) maze 
is cast as an automated planning problem as follows.

\begin{itemize}
    \item The location of the agent is represented by $N^2$ state 
variables $S = \{s_1, \dots, s_{N^2}\}$ where state variable 
$s_i$ represents whether the agent is located at position 
$i\in \{1,\dots,N^2\}$ or not.
    \item The intended movement of the agent is represented by four 
action variables $A = \{a_1, a_2, a_3, a_4\}$ where action 
variables $a_1$, $a_2$, $a_3$ and $a_4$ represent whether the agent 
attempts to move up, down, right or left, respectively.
    \item Mutual exclusion on the intended movement of the agent is 
represented by the global function as follows. 
\begin{align}
    C(\langle s_1, \dots, a_4\rangle) = 
    \begin{cases}
    \mathit{true},& \text{if } a_1 + a_2 + a_3 + a_4 \leq 1\\
    \mathit{false},              & \text{otherwise}
    \end{cases}\nonumber
\end{align}
    \item The initial location of the agent is $s_i = V_i$ for all positions $i\in \{1,\dots,{N^2}\}$.
    \item The final location of the agent is represented by the goal state 
function as follows.
\begin{align}
    G(\langle s_1, \dots, s_{N^2}\rangle) = 
    \begin{cases}
    \mathit{true},& \text{if } s_i = V'_i\\
    & \quad \forall{i\in \{1,\dots,{N^2}\}}\\
    \mathit{false},              & \text{otherwise}
    \end{cases}\nonumber
\end{align}
where $V'_i$ denotes the goal location of the agent (i.e., $V'_i = \mathit{true}$ 
if and only if position $i\in \{1,\dots,{N^2}\}$ is the final location, 
$V'_i = \mathit{false}$ otherwise).
    \item The objective is to minimize total number of intended movements by 
the agent and is represented by the reward function as follows.
\begin{align}
&R(\langle s_1, \dots, a_4\rangle) = a_1 + a_2 + a_3 + a_4\nonumber
\end{align}
    \item The next location of the agent is represented by the state 
transition function $T$ that is a complex function of state and action 
variables $s_1, \dots, s_{N^2}, a_1, \dots, a_4$. The unknown function $T$ 
is approximated by a BNN $\tilde{T}$, and the details of $\tilde{T}$ are 
provided in Table~\ref{tab:bnn}.
\end{itemize}

We submitted problems with $N=3,4,5$ over 
planning horizons $H=4,\dots,10$. Note that this automated planning 
problem is a deterministic version of its original from 
IPPC2011~\cite{Sanner2011}.

\paragraph{Inventory Control} Inventory Control~\cite{Mann2014} is 
the problem of managing inventory of a product with demand cycle length 
$N\in \mathbb{Z}^{+}$, and is cast as an automated planning problem as 
follows.

\begin{itemize}
    \item The inventory level of the product, phase of the demand cycle and 
whether demand is met or not are represented by three state 
variables $S = \{s_1, s_2, s_3\}$ where state variables $s_1$ and $s_2$ have 
non-negative integer domains.
    \item Ordering some fixed amount of inventory is represented by an action 
variable $A = \{a_1\}$.
    \item Meeting the demand is represented by 
the global function as follows.
\begin{align}
    C(\langle s_1, s_2, s_3, a_1\rangle) = 
    \begin{cases}
    \mathit{true},& \text{if } s_3 = \mathit{true}\\
    \mathit{false},              & \text{otherwise}
    \end{cases}\nonumber
\end{align}
    \item The inventory, the phase of the 
demand cycle and meeting the demand are 
set to their initial values $s_i = V_i$ 
for all ${i\in \{1,2,3\}}$.
    \item Meeting the final demand is 
represented by the goal state function as follows.
\begin{align}
    G(\langle s_1, s_2, s_3\rangle) = 
    \begin{cases}
    \mathit{true},& \text{if } s_3 = \mathit{true}\\
    \mathit{false},              & \text{otherwise}
    \end{cases}\nonumber
\end{align}
    \item The objective is to minimize total storage 
cost and is represented by the reward function as follows.
\begin{align}
&R(\langle s_1, s_2, s_3, a_1 \rangle) = c s_1\nonumber
\end{align} 
where $c$ denotes the unit storage cost.
    \item The next inventory level, the next phase of the demand cycle and 
whether the next demand is met or not are represented 
by the state transition function $T$ that is a complex 
function of state and action variables $s_1, s_2, s_3, a_1$. The unknown 
function $T$ is approximated by a BNN $\tilde{T}$, and the details of 
$\tilde{T}$ are provided in Table~\ref{tab:bnn}.
\end{itemize}

We submitted problems with two 
demand cycle lengths $N \in \{2,4\}$ over planning horizons 
$H=5,\dots,8$. The values of parameters are chosen as $m_1 = 4$ 
and $m_2 = 0$.

\paragraph{System Administrator} System 
Administrator~\cite{Guestrin2001,Sanner2011} is the problem of maintaining 
a computer network of size $N$ and is cast as an automated planning 
problem as follows.

\begin{itemize}
    \item The age of computer $i\in \{1,\dots,N\}$, and whether 
computer $i\in \{1,\dots,N\}$ is running or not, are represented 
by $2N$ state variables $S = \{s_1, \dots, s_{2N}\}$ where state 
variables $s_1, \dots, s_{N}$ have non-negative integer domains.
    \item Rebooting computers $i\in \{1,\dots,N\}$ are represented 
by $N$ action variables $A = \{a_1, \dots, a_{N}\}$.
    \item The bounds on the number of computers that can be rebooted 
and the requirement that all computers must be running are 
represented by global function as follows.
\begin{align}
    C(\langle s_1, \dots, a_{N}\rangle) = 
    \begin{cases}
    \mathit{true},& \text{if } \sum_{i=1}^{N} a_{i} \leq a^{max}\\ 
    &\text{and }s_{i} = \mathit{true}\\
    &\quad \forall{i\in \{N+1,\dots,2N\}}\\
    \mathit{false},              & \text{otherwise}
    \end{cases}\nonumber
\end{align}
where $a^{max}$ is the maximum on the number of computers 
that can be rebooted at a given time.
    \item The age of computer $i\in \{1,\dots,N\}$, and 
whether computer $i\in \{1,\dots,N\}$ is running 
or not are set to their initial values $s_{i} = V_{i}$ 
for all ${i\in \{1,\dots,2N\}}$.
    \item The requirement that all computers must be 
running in the end is represented by the goal state function as follows.
\begin{align}
    G(\langle s_1, \dots, s_{2N}\rangle) = 
    \begin{cases}
    \mathit{true},& \text{if } s_{i} = \mathit{true}\\
    &\quad \forall{i\in \{N+1,\dots,2N\}}\\
    \mathit{false},              & \text{otherwise}
    \end{cases}\nonumber
\end{align}
    \item The objective is to minimize total number of reboots 
and is represented by the reward function as follows.
\begin{align}
&R(\langle s_1, \dots, s_{2N}, a_1, \dots, a_{N}\rangle) = 
\sum_{i=1}^{N} a_{i}\nonumber
\end{align}
    \item The next age of computer $i\in \{1,\dots,N\}$ and whether 
computer $i\in \{1,\dots,N\}$ will be running or not, are represented 
by the state transition function $T$ that is a complex function 
of state and action variables $s_1, \dots, s_{2N}, a_1, \dots, a_{N}$.  
The unknown function $T$ is approximated by a BNN $\tilde{T}$, and the 
details of $\tilde{T}$ are provided in Table~\ref{tab:bnn}.
\end{itemize} 

We submitted problems with $N \in \{4,5\}$ computers over planning 
horizons $H=2,3,4$. 
The values of parameters are chosen as $m_1 = 3$ and $m_2 = 0$.

\paragraph{Cellda} Influenced by the famous video 
game~\cite{Nintendo1986}, Cellda~\cite{Say2020a} is the task 
of an agent who must escape from a two dimensional ($N$-by-$N$ where 
$N\in \mathbb{Z}^{+}$) cell through a locked 
door by obtaining the key without getting hit by the enemy, and 
is cast as an automated planning problem as follows.

\begin{itemize}
    \item The location of the agent, the location of the enemy, whether 
the key is obtained or not and whether the agent is alive or not 
are represented by six state variables $S = \{s_1, \dots, s_6\}$ 
where state variables $s_1$ and $s_2$ represent the horizontal and 
vertical locations of the agent, state variables 
$s_3$ and $s_4$ represent the horizontal and vertical locations of the 
enemy, state variable $s_5$ represents whether 
the key is obtained or not, and state variable 
$s_6$ represents whether the agent is alive or not. State variables 
$s_1$, $s_2$, $s_3$ and $s_4$ have positive integer domains.
    \item The intended movement of the agent is represented by four action 
variables $A = \{a_1, a_2, a_3, a_4\}$ where action 
variables $a_1$, $a_2$, $a_3$ and $a_4$ represent whether the agent 
intends to move up, down, right or left, respectively.
    \item Mutual exclusion on the intended movement of the agent, the 
boundaries of the maze and requirement that the agent must be alive are 
represented by global function as follows.
\begin{align}
    C(\langle s_1, \dots, a_4\rangle) = 
    \begin{cases}
    \mathit{true},& \text{if } a_1 + a_2 + a_3 + a_4 \leq 1\\
    &\text{and } 0 \leq s_i < N \quad \forall{i\in \{1,2\}}\\
    &\text{and } s_6 = \mathit{true}\\
    \mathit{false},              & \text{otherwise}
    \end{cases}\nonumber
\end{align}
    \item The location of the agent, the location of the 
enemy, whether the key is obtained or not, and whether the agent 
is alive or not are set to their initial values $s_i = V_i$ for all 
${i\in \{1,\dots,6\}}$.
    \item The goal location of the agent (i.e., the location of the door), 
the requirement that the agent must be alive in the end and the requirement 
that the key must be obtained are represented by the goal state 
function as follows.
\begin{align}
    G(\langle s_1, \dots, s_6\rangle) = 
    \begin{cases}
    \mathit{true},& \text{if } s_1 = V'_1 \text{ and } s_2 = V'_2\\ 
    &\text{and } s_5 = \mathit{true} \text{ and } s_6 = \mathit{true}\\
    \mathit{false},              & \text{otherwise}
    \end{cases}\nonumber
\end{align}
where $V'_1$ and $V'_2$ denote the goal location of the agent (i.e., the 
location of the door).
    \item The objective is to minimize total number of intended movements by 
the agent and is represented by the reward function as follows.
\begin{align}
&R(\langle s_1, \dots, a_4\rangle) = a_1 + a_2 + a_3 + a_4\nonumber
\end{align}
    \item The next location of the agent, the next location of the enemy, 
whether the key will be obtained or not, and whether the agent will 
be alive or not, are represented by the state 
transition function $T$ that is a complex function of state and action 
variables $s_1, \dots, s_6, a_1, \dots, a_4$. The unknown function $T$ 
is approximated by a BNN $\tilde{T}$, and the details of $\tilde{T}$ are 
provided in Table~\ref{tab:bnn}.
\end{itemize}

We submitted problems with maze size $N=4$ over planning horizons 
$H=8,\dots,12$ with two different enemy policies. The values of 
parameters are chosen as $m_1=2$ and $m_2=0$.

\bibliographystyle{IEEEtran}
\bibliography{mybibliography}

\begin{thebibliography}{10}
\providecommand{\url}[1]{#1}
\csname url@samestyle\endcsname
\providecommand{\newblock}{\relax}
\providecommand{\bibinfo}[2]{#2}
\providecommand{\BIBentrySTDinterwordspacing}{\spaceskip=0pt\relax}
\providecommand{\BIBentryALTinterwordstretchfactor}{4}
\providecommand{\BIBentryALTinterwordspacing}{\spaceskip=\fontdimen2\font plus
\BIBentryALTinterwordstretchfactor\fontdimen3\font minus
  \fontdimen4\font\relax}
\providecommand{\BIBforeignlanguage}[2]{{%
\expandafter\ifx\csname l@#1\endcsname\relax
\typeout{** WARNING: IEEEtran.bst: No hyphenation pattern has been}%
\typeout{** loaded for the language `#1'. Using the pattern for}%
\typeout{** the default language instead.}%
\else
\language=\csname l@#1\endcsname
\fi
#2}}
\providecommand{\BIBdecl}{\relax}
\BIBdecl

\bibitem{Nau2004}
D.~Nau, M.~Ghallab, and P.~Traverso, \emph{Automated Planning: Theory \&
  Practice}.\hskip 1em plus 0.5em minus 0.4em\relax San Francisco, CA, USA:
  Morgan Kaufmann Publishers Inc., 2004.

\bibitem{Kautz1992}
H.~Kautz and B.~Selman, ``Planning as satisfiability,'' in \emph{Proceedings of
  the Tenth European Conference on Artificial Intelligence}, ser. ECAI'92,
  1992, pp. 359--363.

\bibitem{Hoffmann2001}
J.~Hoffmann and B.~Nebel, ``The {FF} planning system: Fast plan generation
  through heuristic search,'' in \emph{Journal of Artificial Intelligence
  Research}, vol.~14.\hskip 1em plus 0.5em minus 0.4em\relax USA: AI Access
  Foundation, 2001, pp. 253--302.

\bibitem{Helmert2006}
M.~Helmert, ``The fast downward planning system,'' in \emph{Journal Artificial
  Intelligence Research}, vol.~26.\hskip 1em plus 0.5em minus 0.4em\relax USA:
  AI Access Foundation, 2006, pp. 191--246.

\bibitem{Pommerening2014}
F.~Pommerening, G.~R{\"o}ger, M.~Helmert, and B.~Bonet, ``{LP}-based heuristics
  for cost-optimal planning,'' in \emph{Proceedings of the Twenty-Fourth
  International Conference on Automated Planning and Scheduling}, ser.
  ICAPS’14.\hskip 1em plus 0.5em minus 0.4em\relax AAAI Press, 2014, pp.
  226--234.

\bibitem{Davies2015}
T.~O. Davies, A.~R. Pearce, P.~J. Stuckey, and N.~Lipovetzky, ``Sequencing
  operator counts,'' in \emph{Proceedings of the Twenty-Fifth International
  Conference on Automated Planning and Scheduling}.\hskip 1em plus 0.5em minus
  0.4em\relax AAAI Press, 2015, pp. 61--69.

\bibitem{Shen1989}
W.-M. Shen and H.~A. Simon, ``Rule creation and rule learning through
  environmental exploration,'' in \emph{Proceedings of the Eleventh
  International Joint Conference on Artificial Intelligence}, ser.
  IJCAI’89.\hskip 1em plus 0.5em minus 0.4em\relax San Francisco, CA, USA:
  Morgan Kaufmann Publishers Inc., 1989, pp. 675–--680.

\bibitem{Gil1992}
Y.~Gil, ``Acquiring domain knowledge for planning by experimentation,'' Ph.D.
  dissertation, Carnegie Mellon University, USA, 1992.

\bibitem{Bennett1996}
S.~W. Bennett and G.~F. DeJong, ``Real-world robotics: Learning to plan for
  robust execution,'' in \emph{Machine Learning}, vol.~23, 1996, pp. 121--161.

\bibitem{Benson1997}
S.~S. Benson, ``Learning action models for reactive autonomous agents,'' Ph.D.
  dissertation, Stanford University, Stanford, CA, USA, 1997.

\bibitem{Say2017}
B.~Say, G.~Wu, Y.~Q. Zhou, and S.~Sanner, ``Nonlinear hybrid planning with deep
  net learned transition models and mixed-integer linear programming,'' in
  \emph{Proceedings of the Twenty-Sixth International Joint Conference on
  Artificial Intelligence}, ser. IJCAI'17, 2017, pp. 750--756.

\bibitem{Say2018}
B.~Say and S.~Sanner, ``Planning in factored state and action spaces with
  learned binarized neural network transition models,'' in \emph{Proceedings of
  the Twenty-Seventh International Joint Conference on Artificial
  Intelligence}, ser. IJCAI'18, 2018, pp. 4815--4821.

\bibitem{Say2020c}
B.~Say, ``Optimal planning with learned neural network transition models,''
  Ph.D. dissertation, University of Toronto, Toronto, ON, Canada, 2020.

\bibitem{Say2019}
B.~Say, S.~Sanner, and S.~Thi{\'e}baux, ``Reward potentials for planning with
  learned neural network transition models,'' in \emph{Proceedings of the
  Twenty-Fifth International Conference on Principles and Practice of
  Constraint Programming}, T.~Schiex and S.~de~Givry, Eds.\hskip 1em plus 0.5em
  minus 0.4em\relax Cham: Springer International Publishing, 2019, pp.
  674--689.

\bibitem{Wu2020}
G.~Wu, B.~Say, and S.~Sanner, ``Scalable planning with deep neural network
  learned transition models,'' \emph{Journal of Artificial Intelligence
  Research}, vol.~68, pp. 571--606, 2020.

\bibitem{Say2021}
B.~Say, ``A unified framework for planning with learned neural network
  transition models,'' in \emph{Proceedings of the Thirty-Fifth AAAI Conference
  on Artificial Intelligence}, 2021, pp. 5016--5024.

\bibitem{Say2020a}
B.~Say and S.~Sanner, ``Compact and efficient encodings for planning in
  factored state and action spaces with learned binarized neural network
  transition models,'' \emph{Artificial Intelligence}, vol. 285, p. 103291,
  2020.

\bibitem{Say2020b}
B.~Say, J.~Devriendt, J.~Nordstr{\"{o}}m, and P.~Stuckey, ``Theoretical and
  experimental results for planning with learned binarized neural network
  transition models,'' in \emph{Proceedings of the Twenty-Sixth International
  Conference on Principles and Practice of Constraint Programming}, H.~Simonis,
  Ed.\hskip 1em plus 0.5em minus 0.4em\relax Cham: Springer International
  Publishing, 2020, pp. 917--934.

\bibitem{Hubara2016}
I.~Hubara, M.~Courbariaux, D.~Soudry, R.~El-Yaniv, and Y.~Bengio, ``Binarized
  neural networks,'' in \emph{Proceedings of the Thirtieth International
  Conference on Neural Information Processing Systems}.\hskip 1em plus 0.5em
  minus 0.4em\relax USA: Curran Associates Inc., 2016, pp. 4114--4122.

\bibitem{Ioffe2015}
S.~Ioffe and C.~Szegedy, ``Batch normalization: Accelerating deep network
  training by reducing internal covariate shift,'' in \emph{Proceedings of the
  Thirty-Second International Conference on International Conference on Machine
  Learning}, ser. ICML.\hskip 1em plus 0.5em minus 0.4em\relax JMLR.org, 2015,
  pp. 448--456.

\bibitem{Abio2014}
I.~Ab{\'i}o and P.~Stuckey, ``Encoding linear constraints into {SAT},'' in
  \emph{Principles and Practice of Constraint Programming}.\hskip 1em plus
  0.5em minus 0.4em\relax Springer Int Publishing, 2014, pp. 75--91.

\bibitem{Asin2009}
R.~Asin and R.~Nieuwenhuis, ``Cardinality networks and their applications and
  oliveras, albert and rodriguez-carbonell, enric,'' in \emph{International
  Conference on Theory and Applications of Satisfiability Testing}, 2009, pp.
  167--180.

\bibitem{Sanner2011}
S.~Sanner and S.~Yoon, ``International probabilistic planning competition,''
  2011.

\bibitem{Mann2014}
T.~Mann and S.~Mannor, ``Scaling up approximate value iteration with options:
  Better policies with fewer iterations,'' in \emph{Proceedings of the
  Thirty-First International Conference on Machine Learning}, ser. Machine
  Learning Research, E.~P. Xing and T.~Jebara, Eds., vol.~32.\hskip 1em plus
  0.5em minus 0.4em\relax Bejing, China: PMLR, 2014, pp. 127--135.

\bibitem{Guestrin2001}
C.~Guestrin, D.~Koller, and R.~Parr, ``Max-norm projections for factored
  {MDP}s,'' in \emph{Seventeenth International Joint Conferences on Artificial
  Intelligence}, 2001, pp. 673--680.

\bibitem{Nintendo1986}
Nintendo, ``The legend of zelda,'' 1986.

\end{thebibliography}

\end{document}